\documentclass{llncs}
\usepackage{times}
\usepackage{latexsym}
\usepackage{xcolor}
\usepackage{url}
\usepackage{graphicx}
\usepackage{amsmath}
\usepackage{multirow}
\graphicspath{ {Figure/} }
\usepackage{graphicx}
\usepackage{tabularx}
\usepackage{subcaption}
\newcommand*{\affmark}[1][*]{\textsuperscript{#1}}

\title{Semantic Refinement GRU-based Neural Language Generation for Spoken Dialogue Systems}
\author{%
\textbf{Van-Khanh Tran\affmark[1,2] \and Le-Minh Nguyen\affmark[1]} \\ 
\institute{\affmark[1]Japan Advanced Institute of Science and Technology, JAIST\\
					1-1 Asahidai, Nomi, Ishikawa, 923-1292, Japan\\
\email{\{tvkhanh, nguyenml\}@jaist.ac.jp}\\
\affmark[2]University of Information and Communication Technology, ICTU\\
	Thai Nguyen University, Vietnam\\
\email{tvkhanh@ictu.edu.vn}}%
}

\begin{document}

\maketitle       

\begin{abstract}
Natural language generation (NLG) plays a critical role in spoken dialogue systems. This paper presents a new approach to NLG by using recurrent neural networks (RNN), in which a gating mechanism is applied before RNN computation. This allows the proposed model to generate appropriate sentences. The RNN-based generator can be learned from unaligned data by jointly training sentence planning and surface realization to produce natural language responses. The model was extensively evaluated on four different NLG domains. The results show that the proposed generator achieved better performance on all the NLG domains compared to previous generators.
\end{abstract}

% ============================================================== %
\section{Introduction}\label{sec:introduction}
Natural Language Generation is a critical component in a Spoken Dialogue System (SDS), and its task is to convert a meaning representation produced by the dialogue manager into natural language utterances. Conventional approaches to NLG follow a pipeline which typically breaks down the task into sentence planning and surface realization. Sentence planning decides the order and structure of sentence, which is followed by a surface realization which converts the sentence structure into the final utterance. Previous approaches to NLG still rely on extensive hand-tuning templates and rules that require expert knowledge of linguistic representation. There are some common and widely approaches to solve NLG problems, including rule-based \cite{cheyer2014method}, corpus-based n-gram models \cite{oh2000stochastic}, and a trainable generator \cite{stent2004trainable}. Joint based generators use a two-step pipeline \cite{rieser2010optimising,duvsek2016sequence}; or applying a joint model for both tasks \cite{thwsjy15,wensclstm15}.

Recently, approaches based on recurrent neural networks have shown advantages in solving the NLG tasks. RNN-based models have been used for NLG as a joint training model \cite{thwsjy15,wensclstm15} and an end-to-end training model \cite{wen2016network}. A recurring problem in such systems is requiring datasets annotated for specific dialogue acts\footnote{A combination of an action type and a list of slot-value pairs. \textit{e.g}. \textit{inform(name=`Frances'; area=`City Center')}} (DA). Moreover, previous works may lack ability to handle cases such as the binary slots (\textit{i.e}., \textit{yes} and \textit{no}) and slots that take \textit{don't\_care} value which cannot be directly delexicalized \cite{thwsjy15},  
or to generalize to unseen domains \cite{wensclstm15}. Furthermore, a problem of the current generators is that the generators produce the next token based on the information from the forward context, whereas the sentence may depend on the backward context. As a result, the generators tend to generate nonsensical utterances.
% \cite{thwsjy15} proposed a heuristic gate to ensure that all the attribute-value information was accurately captured during generation. However, this heuristic gate cannot handle cases such as the binary slots (i.e., \textit{yes} and \textit{no}) and slots that take \textit{don't\_care} value which cannot be directly delexicalized. Furthermore, a problem of the current generators is that the generators produce the next token based on the information from the forward context, whereas some sentences may depend on the backward context.
% Moreover, the previous RNN-based models may have lack of consideration about the order of slot-value pairs during generation. For example, given a DA with pattern: \textit{Compare}(name=\textit{A}, property1=\textit{a1}, property2=\textit{a2}, name=\textit{B}, property1=\textit{b1}, property2=\textit{b2}). The pattern for correct utterances can be: [\textit{A-a1-a2, B-b1-b2}], [\textit{A-a2-a1, B-b2-b1}], [\textit{B-b1-b2, A-a1-a2}], [\textit{B-b1-b2, A-a2-a1}]. Therefore, a generated utterance: "The \textit{A} has \textit{a1} and \textit{\textbf{b1}} properties, while the \textit{B} has \textit{\textbf{a2}} and \textit{b2} properties" is an incorrect response, in which \textit{\textbf{b1}} and \textit{\textbf{a2}} properties were generated in wrong oder. This occasionally leads to inadequate sentences.

We propose a statistical NLG based on a gating mechanism on a GRU model, in which the gating mechanism is applied before RNN computation. The proposed model can learn from unaligned data by jointly training the sentence planning and surface realization to generate required sentences. 
We found that the proposed model can produce sentences in a more correct oder than the existing models. The previous RNN-based generators may have lack of consideration about the order of slot-value pairs during generation. For example, given a DA with pattern: \textit{Compare}(name=\textit{A}, property1=\textit{a1}, property2=\textit{a2}, name=\textit{B}, property1=\textit{b1}, property2=\textit{b2}). The pattern for correct utterances can be: [\textit{A-a1-a2, B-b1-b2}], [\textit{A-a2-a1, B-b2-b1}], [\textit{B-b1-b2, A-a1-a2}], [\textit{B-b1-b2, A-a2-a1}]. Therefore, a generated utterance: "The \textit{A} has \textit{a1} and \textit{\textbf{b1}} properties, while the \textit{B} has \textit{\textbf{a2}} and \textit{b2} properties" is an incorrect utterance, in which \textit{\textbf{b1}} and \textit{\textbf{a2}} properties were generated in wrong order. This occasionally leads to inadequate sentences. 

We assessed the proposed generators on varied NLG domains, in which the results showed that our proposed method outperforms the previous methods in terms of BLEU \cite{papineni2002bleu} and slot error rate ERR \cite{wensclstm15} scores. To summary, we make three contributions in this study where we: (i) propose two semantic refinement RNN-based models, in which a gating mechanism is applied before computational RNN to refine the original inputs, (ii) conduct extensively experiments on four NLG domains, and (iii) analyze the effectiveness of the proposed models on ability to handle the undelexicalized tokens, and to generalize to unseen domain when limited amount of in-domain data was fed.

% ============================================================== %
\section{Related Work}\label{sec:relatedwork}
Conventional approaches to NLG traditionally split the task into two subtasks: sentence planning and surface realization. Sentence planning deals with mapping of the input semantic symbols onto a linguistic structure, \textit{e.g}., a tree-like or a template structure. The surface realization then converts the structure into an appropriate sentence \cite{stent2004trainable}. Despite their success and wide use in solving NLG problems, these traditional methods still rely on the handcrafted rule-based generators or rerankers. The authors in \cite{oh2000stochastic} proposed a class-based n-gram language model (LM) generator which can learn to generate the sentences for a given DA and then select the best sentences using a rule-based reranker. Some of the limitation of the class-based LMs were addressed in \cite{ratnaparkhi2000trainable} by proposing a method based on a syntactic dependency tree. A phrase-based generator based on factored LMs was introduced in \cite{mairesse2014stochastic}, which can learn from a semantically aligned corpus. 

Recently, RNNs-based approaches have shown promising performance in the NLG domain. The authors in \cite{karpathy2015deep,vinyals2015show} used RNNs in a multi-modal setting to generate captions for images, while a generator using RNNs to create Chinese poetry was also proposed in \cite{zhang2014chinese}. The authors in \cite{lowe2015incorporating} encoded an unstructured textual knowledge source along with previous responses and context to produce a response for technical support queries.
For task-oriented dialogue systems, a combination of a forward RNN generator, a CNN reranker, and a backward RNN reranker was proposed in \cite{thwsjy15} to generate utterances. A semantically conditioned-based Long Short-Term Memory (LSTM) generator was introduced in \cite{wensclstm15}, which proposed a control ``\textit{reading}" gate to the traditional LSTM cell and can learn the gating mechanism and language model jointly. A recurring problem in such systems is the lack of sufficient domain-specific annotated data. 
%-----------------------------------------%
\section{Recurrent Neural Language Generator}\label{sec:method}
The recurrent language generator proposed in this paper based on a RNN language model \cite{mikolov2010recurrent}, which consists of three layers: an input layer, a hidden layer and an output layer. The input to the network at each time step \textit{t} is a 1-hot encoding $\textbf{w}_{t}$ of a token\footnote{Input texts are delexicalized in which slot values are replaced by its corresponding slot tokens.} $w_{t}$ which is conditioned on a recurrent hidden layer $\textbf{h}_{t}$. The output layer $\textbf{y}_{t}$ represents the probability distribution of the next token given previous token $w_{t}$ and hidden $\textbf{h}_{t}$. We can sample from this conditional distribution to obtain the next token in a generated string, and feed it as the next input to the generator. This process finishes when a stop sign is generated \cite{karpathy2015deep}, or some constraint are reached \cite{zhang2014chinese}. The network can generate a sequence of tokens which can be lexicalized\footnote{The process in which slot token is replaced by its value.} to form the required utterance. Moreover, in order to ensure that the generated utterance represents the intended meaning of the given DA, the generator is further conditioned on a vector \textbf{z}, a 1-hot vector representation of DA. Inspired by work in \cite{wang2016inner}, we propose an intuition: \textit{Gating before computation}, in which we add gating mechanism before the RNN computation to semantically refine the input tokens. The following sections present two proposed Semantic Refinement (SR) gating based RNN generators.
%-----------------------------------------%
\begin{figure}[!ht]
	\centering
    \includegraphics[width=0.84\textwidth, height=7cm]{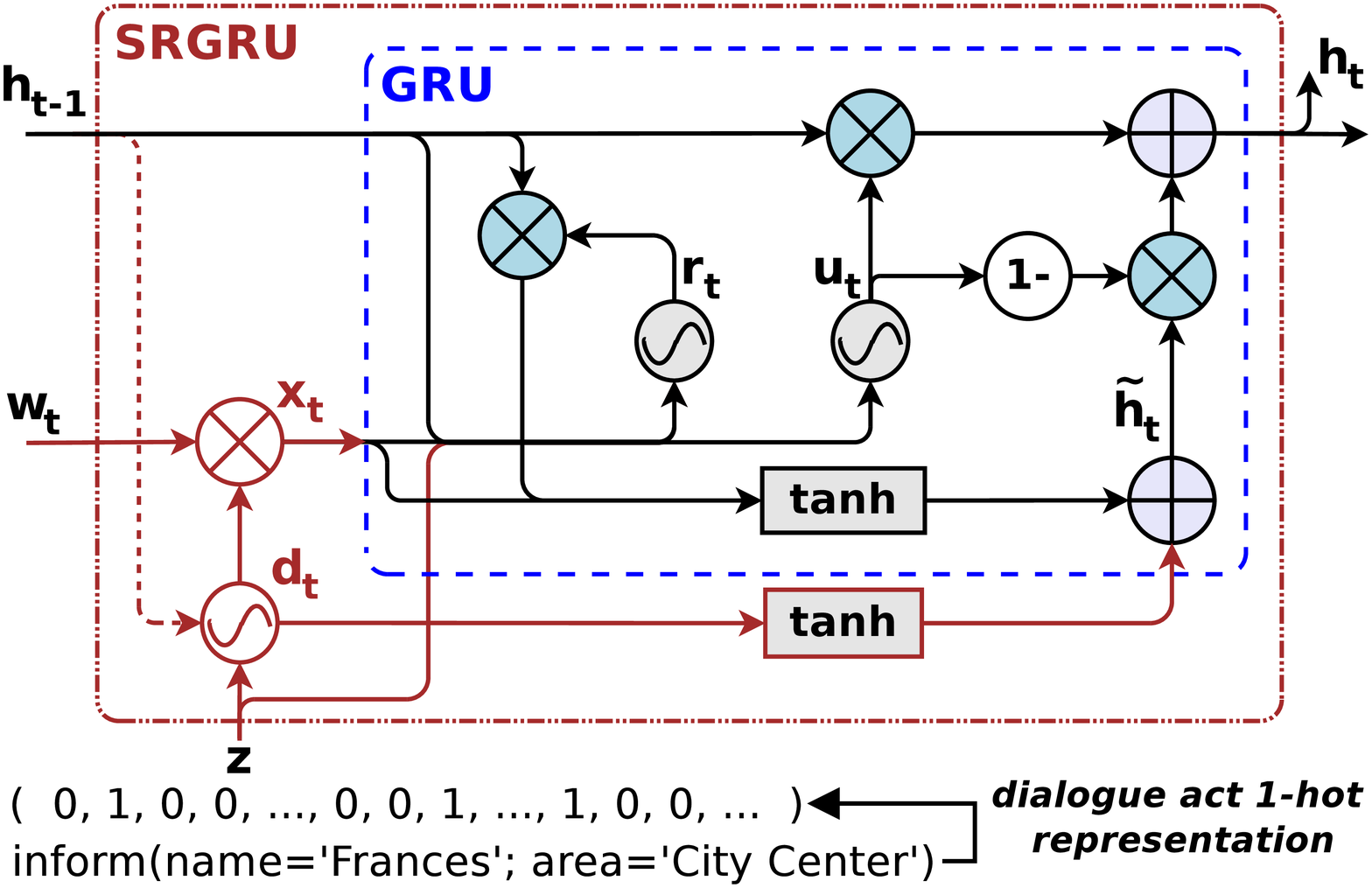}
    \caption{SRGRU-Context cell. The blue dashed box is a traditional GRU cell in charge of surface realization, while the red parts form sentence planning based on a sigmoid control gate $\textbf{d}_{t}$ and a dialogue act $\textbf{z}$. The contextual information $\textbf{h}_{t-1}$ is imported into the refinement gate $\textbf{d}_{t}$ via red dotted line. The SRGRU-Base is achieved by omitting this link.}
    \label{fig:SRGRU-Model}
\end{figure}
%-----------------------------------------------------------%
\subsection{SRGRU-BASE}\label{subsec:srgru-base}
In this model, instead of feeding an input token $\textbf{w}_{t}$ to the RNN model at each time step $t$, the input token is filtered by a semantic gate which is computed as follows:
\begin{equation}\label{eq:d-t-1}
\begin{aligned}
	\textbf{d}_{t}&=\sigma(\textbf{W}_{dz}\textbf{z}) \\
	\textbf{x}_{t}&=\textbf{d}_{t} \odot \textbf{w}_{t} 
\end{aligned}
\end{equation}
where: $\textbf{W}_{dz}$ is a trained matrix to project the given DA representation into the word embedding space, and $\textbf{x}_{t}$ is new input. Here $\textbf{W}_{dz}$ plays a role of sentence planning since it can directly capture which DA features are useful during the generation to encode the input information. The $\odot$ element-wise multiplication plays a part in word-level matching which learns not only the vector similarity, but also preserve information about the two vectors. $\textbf{d}_{t}$ is called a \textit{refinement} gate since the input tokens are refined by the DA information. As a result, we can represent the whole input sentence based on these refined inputs using RNN model. 

In this study, we use GRU, which was recently proposed in \cite{bahdanau2014neural}, instead of LSTM as building computational block for RNN, which is formulated as follows:\vspace{25pt}
\begin{equation}\label{eq:r-t-0}
\textbf{r}_{t}=\sigma(\textbf{W}_{rx}\textbf{x}_{t}+\textbf{W}_{rh}\textbf{h}_{t-1})
\end{equation}
\begin{equation}\label{eq:u-t-0}
\textbf{u}_{t}=\sigma(\textbf{W}_{ux}\textbf{x}_{t}+\textbf{W}_{uh}\textbf{h}_{t-1})
\end{equation}
\begin{equation}\label{eq:h-t-0}
\tilde{\textbf{h}_{t}}=\tanh(\textbf{W}_{hx}\textbf{x}_{t}+r_{t}\odot \textbf{W}_{hh}\textbf{h}_{t-1})
\end{equation}
\begin{equation}
\textbf{h}_{t}= \textbf{u}_{t} \odot \textbf{h}_{t-1} + (1-\textbf{u}_{t}) \odot \tilde{\textbf{h}_{t}}
\end{equation}
where: $\textbf{W}_{rx}, \textbf{W}_{rh}, \textbf{W}_{ux}, \textbf{W}_{uh}, \textbf{W}_{hx}, \textbf{W}_{hh}$ are weight matrices; $\textbf{r}_{t}$, $\textbf{u}_{t}$ are reset and update gate, respectively, and $\odot$ denotes for element-wise product. The Semantic Refinement GRU (SRGRU)-Base architecture is shown in Figure \ref{fig:SRGRU-Model}.

Finally, the output distribution of each token is defined by applying a softmax function $g$ as follows:
\begin{equation}\label{eq:p-t-1}
P(w_{t+1}\mid w_{t}, w_{t-1},...w_{0},\textbf{z}) = g(\textbf{W}_{ho}\textbf{h}_{t})
\end{equation}
where: $\textbf{W}_{ho}$ is learned linear projection matrix. At training time, we use the ground truth token for the previous time step in place of the predicted output. At test time, we implement a simple beam search to over-generate several candidate responses. 

\subsection{SRGRU-CONTEXT}\label{subsec:srgru-context}
SRGRU-Base uses only the DA information to gate the input sequence token by token. As a results, this gating mechanism may not capture the relationship between multiple words. In order to import context information into the gating mechanism, the Equation \ref{eq:d-t-1} is modified as follows:
\begin{equation}\label{eq:d-t-2}
\begin{aligned}
	\textbf{d}_{t}&=\sigma(\textbf{W}_{dz}\textbf{z} + \textbf{W}_{dh}\textbf{h}_{t-1})\\
	\textbf{x}_{t}&=\textbf{d}_{t} \odot \textbf{w}_{t} 
\end{aligned}
\end{equation}
where: $\textbf{W}_{dz}$ and $\textbf{W}_{dh}$ are weight matrices. $\textbf{W}_{dh}$ acts like a key phrase detector that learns to capture the pattern of generation tokens or the relationship between multiple tokens. In other words, the new input $\textbf{x}_{t}$ consists of information of the original input token $\textbf{w}_{t}$, the dialogue act \textbf{z}, and the hidden context $\textbf{h}_{t-1}$. $\textbf{d}_{t}$ is called the \textit{refinement} gate because the input tokens are refined by a combination gating information of the dialogue act \textbf{z} and the previous hidden state $\textbf{h}_{t-1}$. By taking advantage of gating mechanism from the LSTM model \cite{hochreiter1997long} in which the gating mechanism is employed to solve the gradient vanishing and exploding problem, we propose to apply the refinement gate deeper into the GRU cell. Firstly, the GRU reset and update gates can be further influenced on the given dialogue act \textbf{z} and the refined input $\textbf{x}_{t}$. The Equations \eqref{eq:r-t-0} and \eqref{eq:u-t-0} are modified as follows:
\begin{equation}\label{eq:r-t-1} 
\textbf{r}_{t}=\sigma ({\textbf{W}_{rx}\textbf{x}_{t}+\textbf{W}_{rh}\textbf{h}_{t-1}+\textbf{W}_{rz}\textbf{z}})
\end{equation}
\begin{equation}\label{eq:u-t-1}
\textbf{u}_{t}=\sigma (\textbf{W}_{ux}\textbf{x}_{t}+\textbf{W}_{uh}\textbf{h}_{t-1}+\textbf{W}_{uz}\textbf{z})
\end{equation}
where: $\textbf{W}_{rz}$ and $\textbf{W}_{uz}$ act like background detectors that learn to control the style of the generating sentence. Secondly, Equation \eqref{eq:h-t-0} is modified so that the candidate activation $\tilde{\textbf{h}_{t}}$ also depends on the refinement gate,
\begin{equation}\label{eq:h-t-2}
\tilde{\textbf{h}_{t}}=\tanh(\textbf{W}_{hx}\textbf{x}_{t}+\textbf{r}_{t}\odot \textbf{W}_{hh}\textbf{h}_{t-1}) + \tanh(\textbf{W}_{dc}\textbf{d}_{t})
\end{equation}
By this way, the reset and update gates learn not only the long-term dependency but also the gating information from the dialogue act and the previous hidden state. We call the resulting architecture Semantic Refinement GRU (SRGRU)-Context which is shown in Figure \ref{fig:SRGRU-Model}.

%-----------------------------------------%
\subsection{Training}\label{subsec:training}
The cost function was the negative log-likelihood and computed by:
\begin{equation}\label{eq:c-f-1}
F(\theta) = -\sum_{t=1}^{T}\textbf{y}_{t}^{\top}\log{\textbf{p}_{t}}
\end{equation}
where $\textbf{y}_{t}$ is the ground truth word distribution, $\textbf{p}_{t}$ is the predicted word distribution, $T$ is length of the input sequence. 
The generators were trained by treating each sentence as a mini-batch with the \textit{$l_{2}$} regularization was added to the cost function for every 10 training examples. The models were initialized with pre-trained word vectors GLOVE \cite{pennington2014glove} and optimized by using stochastic gradient descent and back propagation through time. To prevent over-fitting, early stopping was implemented using a validation set.
%-----------------------------------------%
\subsection{Decoding}\label{subsec:decoding}
The decoding phase we employ here is similar to \cite{tran2017natural} which consists of over-generation and re-ranking phases. The forward generator, in the over-generation phase, is conditioned on the given DA uses a beam search algorithm to generate candidate utterances, whereas the cost of forward generator $F_{fw}(\theta)$, in the re-ranking phase, is computed to form the re-ranking score $R$ as follows:
\begin{equation}\label{eq:r-score-1}
R = F_{fw}(\theta) + \lambda ERR
\end{equation}
where $\lambda$ is a trade off constant which is set to a large value in order to severely penalize nonsensical outputs. The slot error rate $ERR$, which is the number of generated slots that are either redundant or missing, is computed by:
\begin{equation}
ERR = \frac{p + q}{N}
\end{equation}
where $N$ is the total number of slots in DA, and $p$, $q$ is the number of missing and redundant slots, respectively. The ERR re-ranking criteria as mentioned in \cite{wensclstm15} cannot handle arbitrary slot-value pairs, \textit{i.e.} \textit{binary} slots or slots that take \textit{don’t\_care} value, because such these pairs cannot be delexicalized and matched.

% ============================================================== %
\section{Experiments}\label{sec:experiments}
\subsection{Datasets}\label{subsec:datasets}
We conducted experiments using four different NLG domains: finding a restaurant, finding a hotel, buying a laptop, and buying a television. The Restaurant and Hotel domains were collected in \cite{wensclstm15} which contain system dialogue acts, shared slots, and specific domain slots. The Laptop and TV datasets have been released in \cite{wen2016multi} with about 13K distinct DAs in the Laptop and 7K distinct DAs in the TV. These two datasets have a much larger input space but only one training example for each DA so that the system must learn partial realization of concepts and be able to recombine and apply them to unseen DAs. The number of dialogue act types and slots of datasets is also larger than in Restaurant and Hotel datasets. As a result, the NLG tasks for the Laptop and TV datasets become much harder. 
%-----------------------------------------%
\subsection{Experimental Setups}\label{subsec:experimental-setups}

The generators were implemented using the TensorFlow library \cite{abadi2016tensorflow} and trained by partitioning each of the datasets into training, validation and testing set in the ratio 3:1:1. The hidden layer size was set to be 80, and the generators were trained with a $70\%$ of dropout rate. We perform 5 runs with different random initialization of the network and the training is terminated by using early stopping as described in Section \ref{subsec:training}. We select model that yields the highest BLEU score on the validation set. The decoder procedure used beam search with a beam width of 10. We set $\lambda$ to 1000 to severely discourage the reranker from selecting utterances which contain either redundant or missing slots. For each DA, we over-generated 20 candidate utterances and selected the top 5 realizations after reranking. Because the proposed models work stochastically, except the results reported in Table \ref{tab:tab-performance}, all the results shown were averaged over 5 randomly initialized networks.

Since some sentences may depend on both the past and the future during generation, we train another backward SRGRU-Context to utilizing the flexibility of the refinement gate $\textbf{d}_{t}$, in which we tie its weight matrices such $\textbf{W}_{dz}$ and $\textbf{W}_{dh}$ (Equation \ref{eq:d-t-2}) for both. We found that by tying matrix $\textbf{W}_{dz}$ for both forward and backward RNNs, the proposed generator seems to produce more correct and grammatical utterances than those having the only forward RNN. This model called Tying Backward SRGRU-Context (TB-SRGRU). 

In order to better understand the effectiveness of the proposed model, we conduct more experiments to compare the SRGRU-Context with the previous generator SCLSTM in a variety of setups on proportion of training corpus, beam size, and top-$k$ best results. Firstly, the Restaurant and TV datasets were chosen, in which the SCLSTM model obtained the best performances and the NLG task comes from a limited domain to a more diverse domain as described in Section \ref{subsec:datasets}. In this setup, the models were run with different size of training corpus. Secondly, we examined the stability of SRGRU-Context model on different setups of beam size and top-$k$ best results. 
%-----------------------------------------%
\subsection{Evaluation Metrics and Baselines}\label{subsec:evaluation-metrics}
The generator performance was assessed by using two objective evaluation metrics, the BLEU score and the slot error rate ERR. Note that the slot error rate ERR was computed as an auxiliary metric alongside the BLEU score and calculated by averaging slot errors over each of the top 5 realizations in the entire corpus. Both metrics were computed by adopting code from an open source benchmark toolkit for Natural Language Generation\footnotemark[\value{footnote}].

%-----------------------------------------%
We compared our proposed models against with the general GRU (GRU-Base) and three strong baselines released from the NLG toolkit:
\footnotetext{https://github.com/shawnwun/RNNLG}
\begin{itemize}
	\item ENCDEC proposed in \cite{wentoward} which applies the attention mechanism to an RNN encoder-decoder.
    \item HLSTM proposed in \cite{thwsjy15} which uses a heuristic gate to ensure that all of the attribute-value information was accurately captured when generating.
	\item SCLSTM proposed in \cite{wensclstm15} which can learn the gating signal and language model jointly.
\end{itemize} 

\section{Results and Analysis}\label{sec:resultsandanalysis}
% ============================================================== %
\begin{table*}[!ht]
\centering
\caption{Comparison performance on four datasets in terms of the BLEU and the error rate ERR(\%) scores; \textbf{bold} denotes the best and \textbf{\textit{italic}} shows the second best model. The results were produced by training each network on 5 random initialization and selected model with the highest validation BLEU score.}
\label{tab:tab-performance}
%\resizebox{\textwidth}{!}{%
\begin{tabular}{ccccccccc}
\hline
\multirow{2}{*}{Model} & \multicolumn{2}{c}{\textbf{Restaurant}} & \multicolumn{2}{c}{\textbf{Hotel}} & \multicolumn{2}{c}{\textbf{Laptop}} & \multicolumn{2}{c}{\textbf{TV}} \\ \cline{2-9} 
 & BLEU & ERR & BLEU & ERR & BLEU & ERR & BLEU & ERR \\ \hline
ENCDEC & 0.7398 & 2.78\% & 0.8549 & 4.69\% & 0.5108 & 4.04\% & 0.5182 & 3.18\% \\
HLSTM & 0.7466 & 0.74\% & 0.8504 & 2.67\% & 0.5134 & 1.10\% & 0.5250 & 2.50\% \\
SCLSTM & 0.7525 & \textbf{0.38\%} & 0.8482 & 3.07\% & 0.5116 & \textbf{0.79\%} & 0.5265 & 2.31\% \\ \hline \hline
GRU-Base & 0.7381 & 1.41\% & 0.8455 & 2.66\% & 0.5153 & 1.77\% & 0.5245 & 2.03\% \\
SRGRU-Base & 0.7549 & 0.56\% & 0.8640 & \textbf{\textit{1.21}}\% & 0.5190 & 1.56\% & 0.5305 & 1.62\% \\
SRGRU-Context & \textit{\textbf{0.7634}} & 0.49\% & \textbf{0.8776} & \textbf{0.98}\% & \textit{\textbf{0.5191}} & 1.19\% & \textit{\textbf{0.5311}} & \textit{\textbf{1.33\%}} \\
TB-SRGRU & \textbf{0.7637} & \textbf{\textit{0.47}}\% & \textbf{\textit{0.8642}} & 1.56\% & \textbf{0.5208} & \textbf{\textit{0.93}}\% & \textbf{0.5312} & \textbf{1.01\%} \\ \hline
\end{tabular}
%}
\end{table*}
% ============================================================== %
\begin{table*}[!ht]
\centering
\caption{Comparison performance on variety of SRGRU models on four datasets in terms of the BLEU and the error rate ERR(\%) scores. The results were averaged over 5 randomly initialized networks for each proposed model. \textbf{$^{\natural}$} reported in \cite{wensclstm15}.}
\label{tab:tab-average-performance}
%\resizebox{\textwidth}{!}{%
\begin{tabular}{ccccccccc}
\hline
\multirow{2}{*}{Model} & \multicolumn{2}{c}{\textbf{Restaurant}} & \multicolumn{2}{c}{\textbf{Hotel}} & \multicolumn{2}{c}{\textbf{Laptop}} & \multicolumn{2}{c}{\textbf{TV}} \\ \cline{2-9} 
 & BLEU & ERR & BLEU & ERR & BLEU & ERR & BLEU & ERR \\ \hline
SCLSTM\textbf{$^{\natural}$} & 0.7211 & 0.62\% & 0.8020 & 0.78\%  & - & -& - & - \\ 
+deep\textbf{$^{\natural}$} & 0.7310 & \textbf{0.46}\% & 0.8320 & \textbf{0.41}\%  & - & -& - & -\\ \hline
GRU-Base & 0.7208 & 1.55\% & 0.8426 & 1.97\% & 0.5158 & 1.94\% & 0.5244 & 2.11\% \\
SRGRU-Base & 0.7526 & 1.33\% & 0.8622 & 1.12\% & 0.5165 & 1.79\% & 0.5311 & 1.56\% \\
SRGRU-Context & \textbf{0.7614} & 0.99\% & \textbf{0.8677} & 1.75\% & 0.5182 & 1.41\% & 0.5312 & 1.37\% \\
TB-SRGRU & 0.7608 & 0.88\% & 0.8584 & 1.63\% & \textbf{0.5188} & \textbf{1.35}\% & \textbf{0.5316} & \textbf{1.27}\% \\ \hline
\end{tabular}
%}
\end{table*}
Overall, the proposed models SRGRUs consistently achieve better performance in term of the BLEU score in all domains. Especially, on the Hotel and TV datasets, the proposed models outperform the previous methods in both evaluation metrics. Moreover, our models also outperform the GRU basic model (GRU-Base) in all cases. However, the proposed models get worse on the Restaurant and Hotel datasets in terms of the error rate ERR score in comparison with SCLSTM. This indicates the advantage of the proposed refinement gate. A comparison of the two proposed generators is shown in Table \ref{tab:tab-average-performance}: Without the backward RNN reranker, the generator tends to make semantic errors since it gains the higher slot error rate ERR. However, using the backward SRGRU reranker can improve the results in both evaluation metrics. This reranker provides benefit to the generator on producing higher-quality utterances.
%-----------------------------------------%
\begin{figure}
	\centering
    \begin{minipage}{0.5\textwidth}
      \centering
      \includegraphics[width=\textwidth, height=4.7cm]{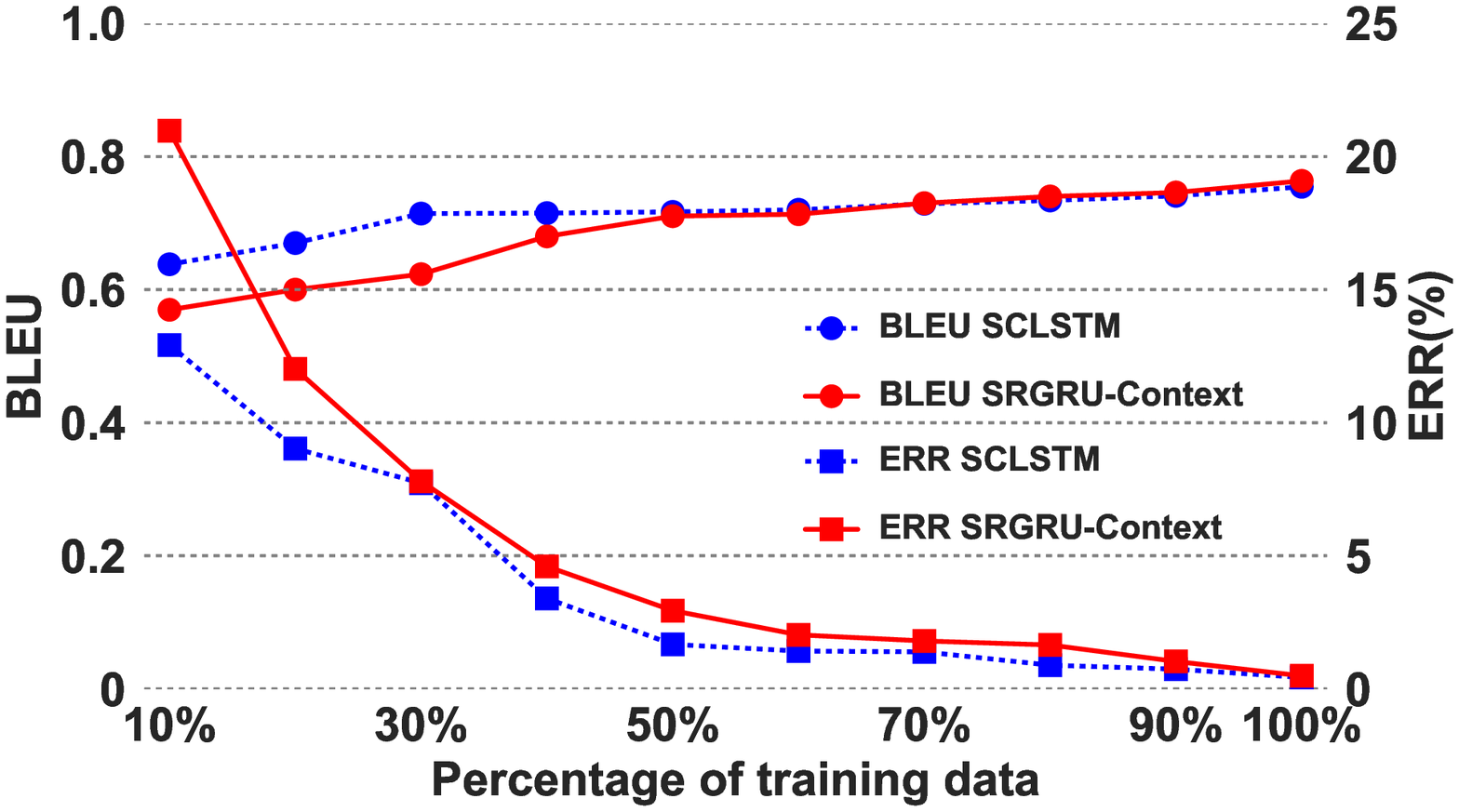}
      \subcaption{Curves on the Restaurant dataset}
      \label{fig:Res-BLEUvsERR}
	\end{minipage}\hfill
    \begin{minipage}{0.5\textwidth}
      \centering
      \includegraphics[width=\textwidth, height=4.7cm]{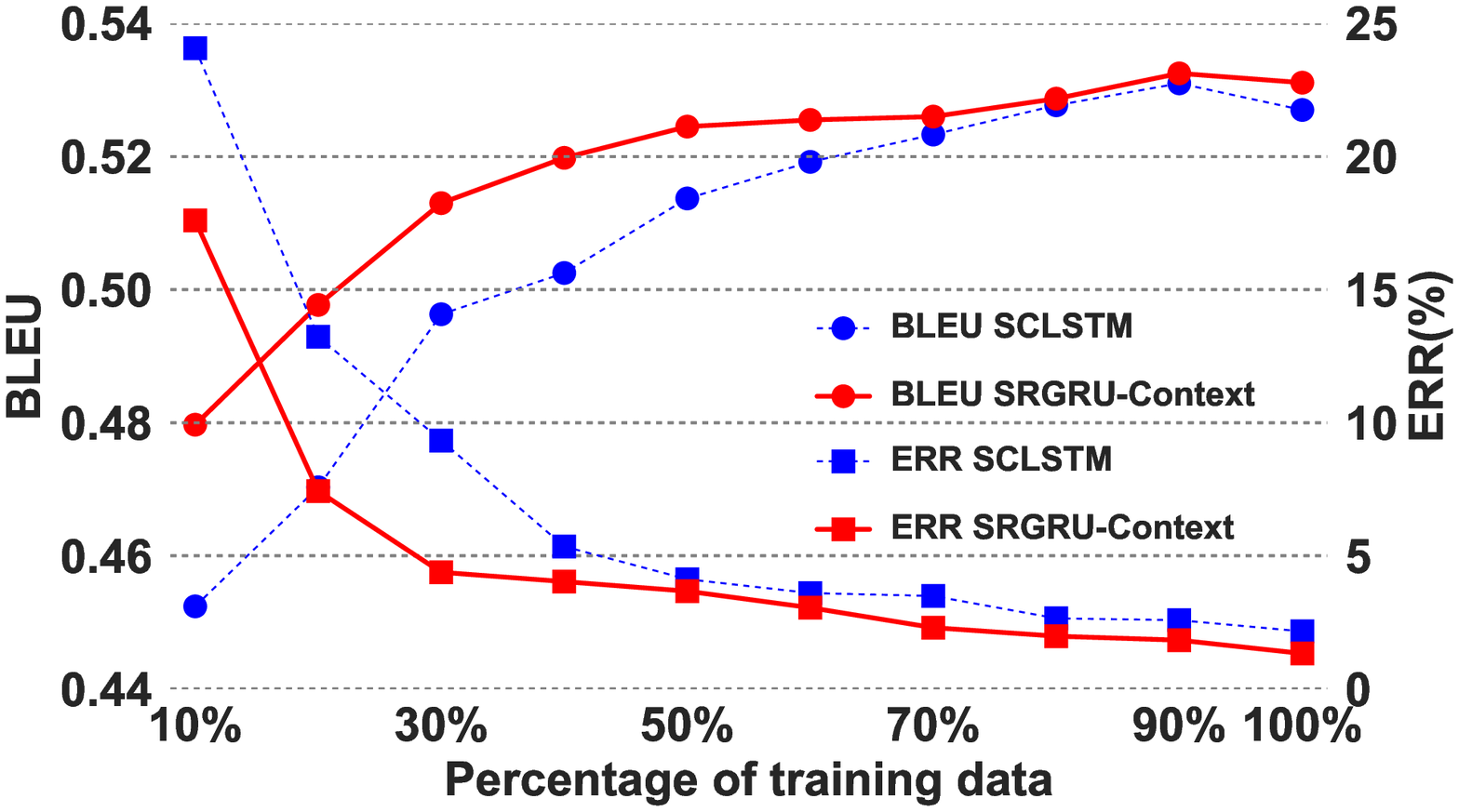}
      \subcaption{Curves on the TV dataset}
      \label{fig:TV-BLEUvsERR} 
    \end{minipage}
\caption{Comparison of two generators SRGRU-Context and SCLSTM which are trained with different proportion of training data.}   
\label{fig:comparison-2models}
\end{figure}
%-----------------------------------------%

Figure \ref{fig:comparison-2models} compares two generators trained with different proportion of data evaluated on two metrics. As can be seen in Figure \ref{fig:Res-BLEUvsERR}, the SCLSTM model achieves better results than SRGRU-Context model on both of BLEU and ERR scores since a small amount of training data was provided. However, the SRGRU-Context obtains the higher BLEU score and slightly higher ERR score as more training data was fed. On the other hand, in a more diverse dataset TV, the SRGRU-Context model consistently outperforms the SCLSTM on both evaluation metrics no matter how much training data is (Figure \ref{fig:TV-BLEUvsERR}). This is mainly due to the ability of refinement gate which feeds to the GRU model a new input $\textbf{x}_{t}$ conveying useful information filtered from the original input and the gating mechanism. Moreover, this gate also keeps the pattern of the generated utterance during generation. As a result, it can have a better realization of unseen slot-value pairs.

\begin{figure}
	\centering
    \begin{minipage}{0.5\textwidth}
      \centering
      \includegraphics[width=\textwidth, height=4.7cm]{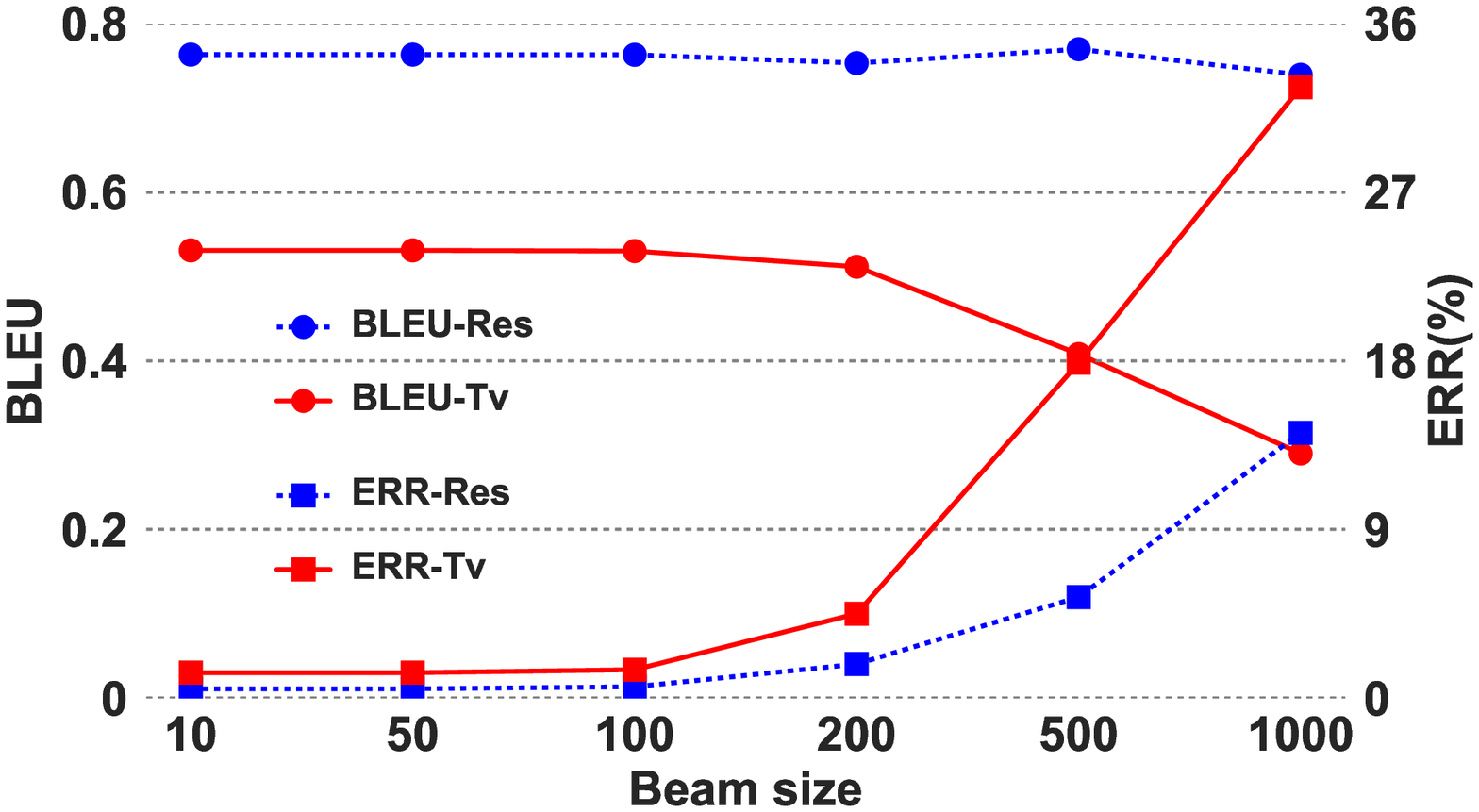}
      \subcaption{Curves on variation of Beam size}
      \label{fig:BEAM-RESvsTV}
	\end{minipage}\hfill
    \begin{minipage}{0.5\textwidth}
      \centering
      \includegraphics[width=\textwidth, height=4.7cm]{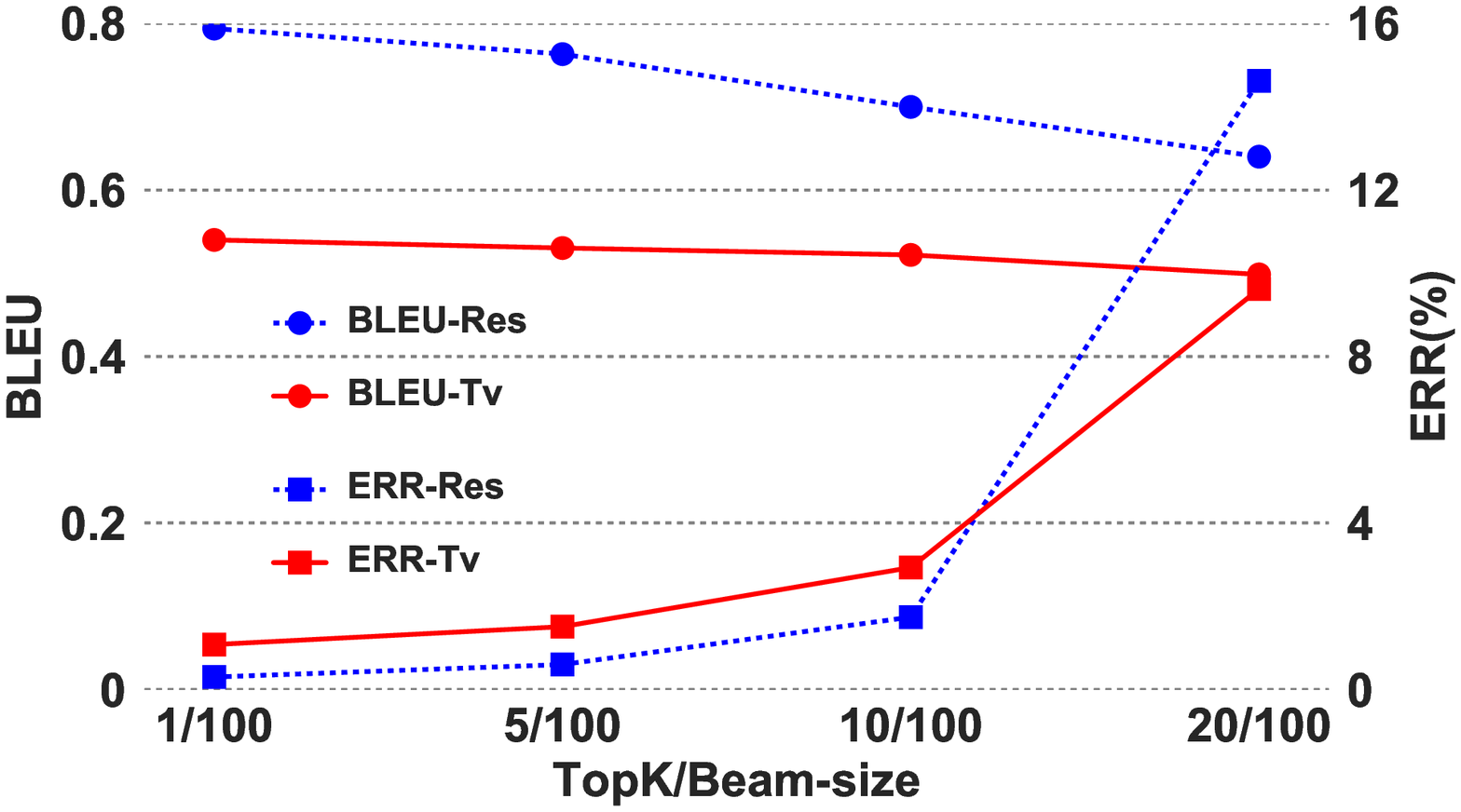}
      \subcaption{Curves on variation of Top-$k$/Beam-size}
      \label{fig:TopK_RESvsTV} 
    \end{minipage}
\caption{RGRU-Context generator was trained with different Beam size (\textit{a}) and Top-$k$ best results (\textit{b}) and evaluated on Restaurant and TV datasets.}   
\label{fig:BEAM-TopK-RESvsTV}
\end{figure}

Figure \ref{fig:BEAM-RESvsTV} shows an effect of beam size on the SRGRU-Context model evaluated on Restaurant and TV datasets. As can be seen that, the model performs worse in terms of degrading the BLEU score and upgrading the slot error rate ERR when the beam size increases. The model seems to perform best with beam size less than $100$. Figure \ref{fig:TopK_RESvsTV} presents an effect of top-$k$ best results in which we fixed the beam size at $100$ and top-$k$ best results varied as $k$ = $1$, $5$, $10$ and $20$. In each case, the BLEU and the error rate ERR scores were computed on Restaurant and TV datasets. The results are consistent with Figure \ref{fig:BEAM-RESvsTV} in which the BLEU and ERR scores get worse as more top-$k$ best utterances were chosen.

%-----------------------------------------%

% ============================================================== %
\begin{table}[ht]
\centering
\caption{Comparison of top responses generated for some input dialogue acts between different models. Errors are marked in color (\textcolor{red}{missing}, \textcolor{blue}{misplaced}, \textcolor{orange}{repeated}, \textcolor{green}{grammar} information). \textbf{$^{\dag}$} and \textbf{$^{\natural}$} denotes the baselines and the proposed models, respectively.}

\label{tab:comparison}
\resizebox{\textwidth}{!}{%
\begin{tabularx}{1.24\textwidth}{lX}
 \textbf{Model} & \textbf{Generated Responses} \\ \hline 
%%%%%%%%%%%%%%%%%%%%%%%%%%%%%%%%%%% Restaurant Example %%%%%%%%%%%%%%%%%%%%%%%%%%
 \textbf{\textit{Input DA}} & \textit{inform\_count(type=restaurant; count=2; food=Basque; kidsallowed=no; pricerange=moderate)} \\
\textbf{\textit{Reference}} & \textit{There are 2 restaurants where no children are allowed in the moderate price range and serving Basque food.} \\
 ENCDEC\textbf{$^{\dag}$} & There are 2 restaurants that are moderately priced and do not allow kids. [\textcolor{red}{Basque}] \\
 HLSTM\textbf{$^{\dag}$} & There are 2 Basque restaurants that are moderately priced and \textcolor{green}{does not} allow kids. \\
 SCLSTM\textbf{$^{\dag}$} & There are 2 Basque restaurants that are moderate and \textcolor{green}{does not} allow children. \\
SRGRU-C\textbf{$^{\natural}$} & There are 2 moderate restaurants that serve Basque food and \textcolor{green}{does not} allow children. \\
TB-SRGRU\textbf{$^{\natural}$} & There are 2 moderately priced Basque restaurants that do not allow kids \\ \hline
%%%%%%%%%%%%%%%%%%%%%%%%%%%%%%% Hotel Example %%%%%%%%%%%%%%%%%%%%%%%%%%%%%%
 \textbf{\textit{Input DA}} & \textit{inform(name=the Carriage Inn; hasinternet=yes; dogsallowed=no)} \\
\textbf{\textit{Reference}} & \textit{The Carriage Inn is a nice place, it has internet and where no dogs are allowed}. \\
ENCDEC\textbf{$^{\dag}$} & The Carriage Inn does not allow dogs. [\textcolor{red}{has internet}] \\
HLSTM\textbf{$^{\dag}$} & The Carriage Inn does not allow dogs, \textcolor{green}{do not} allow dogs [\textcolor{orange}{do not allow dogs}]. \\
SCLSTM\textbf{$^{\dag}$} & The Carriage Inn does not allow dogs and has internet. \\
SRGRU-C\textbf{$^{\natural}$} & The Carriage Inn has internet and does not allow dogs.\\
TB-SRGRU\textbf{$^{\natural}$} & The Carriage Inn does not allow dogs and it has internet. \\ \hline
%%%%%%%%%%%%%%%%%%%%%%%%%%%%%%% TV Example %%%%%%%%%%%%%%%%%%%%%%%%%%%%%%
\textbf{\textit{Input DA}} & \textit{compare(name=Triton 52; ecorating=A+; family=L7; name=Hades 76; ecorating=C; family=L9)}
 \\
 \textbf{\textit{Reference}} & \textit{Compared to Triton 52 which is in the A+ eco rating and is in the L7 product family, Hades 76 is in the C eco rating and is in the L9 product family. Which one do you prefer?}
 \\
ENCDEC\textbf{$^{\dag}$} & The Triton 52 has an A+ eco rating, the Hades 76 in the L7 product family and has \textcolor{green}{an} C eco rating. [\textcolor{blue}{L7}, \textcolor{red}{L9}] \\
HLSTM\textbf{$^{\dag}$} & The Triton 52 is in the L7 product family with an A+ eco rating, while the Hades 76 has a C eco rating, which do you prefer? [\textcolor{red}{L9}] \\
SCLSTM\textbf{$^{\dag}$} & The Triton 52 has an A+ eco rating, the Hades 76 is in the L7 family and has \textcolor{green}{a} eco rating of C. [\textcolor{blue}{L7}, \textcolor{red}{L9}]\\
SRGRU-C\textbf{$^{\natural}$} & The Triton 52 is in the L7 product family and an A+ eco rating, the Hades 76 is in the L9 family and has an C eco rating.\\
TB-SRGRU\textbf{$^{\natural}$} & The Triton 52 has an A+ eco rating, in the L7 product family, the Hades 76 has a C eco rating and is in the L9 product family.
\end{tabularx}%
}
\end{table}

Table \ref{tab:comparison} shows comparison of top responses generated by different models for given DAs. Firstly, both models SCLSTM and SRGRU-Context seem to produce the same kind of error, for example, \textit{grammar} mistakes or \textit{missing} information, partly because of using the same idea about gating mechanism. However, TB-SRGRU, with tying the refinement gate, has ability to fix this problem and produce the correct utterances (row 1 of Table \ref{tab:comparison}). Secondly, as noted earlier, one problem of the previous methods is the ability to handle the \textit{binary} slot and slots that take \textit{don't\_care} value. Both SCLSTM and the proposed models are able to handle this problem (row 2 of Table \ref{tab:comparison}). Finally, the TV dataset is more diverse and much harder than the others because the order of slot-value pairs should be considered during generation. 
For example, to generate a comparison sentence of 2 items \textit{Triton 52} and \textit{Hades 76} for given dialogue act \textit{compare(name=Triton 52; ecorating=A+; family=L7; name=Hades 76; ecorating=C; family=L9)}, the generator should consider that \textit{A+} and \textit{L7} values belong to the former item while \textit{C} and \textit{L9} values to the latter. 
The HLSTM tends to make the \textit{repeated} information error while the ENCDEC and SCLSTM seem to \textit{misplace} the slot value during generation. Take \textit{L7} value, for instance, which should be generated follow the \textit{Triton 52} instead of \textit{Hades 76} as in row 3 of Table \ref{tab:comparison}. We found that both proposed models SRGRU-Context and TB-SRGRU can deal with this problem to generate appropriate utterances.

\section{Conclusion and Future Work}\label{sec:conclusion}
We propose a gating mechanism GRU-based generator, in which we introduced a refinement gate to semantically refine the original input tokens. The refined inputs conveying meaningful information are then fed into the GRU cell. The proposed models can learn from the unaligned data to produce natural language responses conditioned on the given DA. We extensively evaluated our model on four NLG datasets and compared against the previous generators. The results show that the proposed models obtain better performance than the existing generators on all of four NLG domains in terms of the BLEU and ERR metrics. In the future, we plan to further investigate the gating mechanism to multi-domain NLG since the refinement gate shows its ability to handle the unseen slot-value pairs.

\section*{Acknowledgment}
This work was supported by the JSPS KAKENHI Grant number JP15K16048.
\bibliography{pacling2017}

\begin{thebibliography}{10}
\providecommand{\url}[1]{\texttt{#1}}
\providecommand{\urlprefix}{URL }

\bibitem{abadi2016tensorflow}
Abadi, M., Agarwal, A., Barham, P., Brevdo, E., Chen, Z., Citro, C., Corrado,
  G.S., Davis, A., Dean, J., Devin, M., et~al.: Tensorflow: Large-scale machine
  learning on heterogeneous distributed systems. arXiv preprint
  arXiv:1603.04467  (2016)

\bibitem{bahdanau2014neural}
Bahdanau, D., Cho, K., Bengio, Y.: Neural machine translation by jointly
  learning to align and translate. arXiv preprint arXiv:1409.0473  (2014)

\bibitem{cheyer2014method}
Cheyer, A., Guzzoni, D.: Method and apparatus for building an intelligent
  automated assistant (Mar~18 2014), uS Patent 8,677,377

\bibitem{duvsek2016sequence}
Du{\v{s}}ek, O., Jur{\v{c}}{\'\i}{\v{c}}ek, F.: Sequence-to-sequence generation
  for spoken dialogue via deep syntax trees and strings. arXiv preprint
  arXiv:1606.05491  (2016)

\bibitem{hochreiter1997long}
Hochreiter, S., Schmidhuber, J.: Long short-term memory. Neural computation
  (1997)

\bibitem{karpathy2015deep}
Karpathy, A., Fei-Fei, L.: Deep visual-semantic alignments for generating image
  descriptions. In: Proc. CVPR. pp. 3128--3137 (2015)

\bibitem{lowe2015incorporating}
Lowe, R., Pow, N., Serban, I., Charlin, L., Pineau, J.: Incorporating
  unstructured textual knowledge sources into neural dialogue systems. In: NIPS
  Workshop MLNLU (2015)

\bibitem{mairesse2014stochastic}
Mairesse, F., Young, S.: Stochastic language generation in dialogue using
  factored language models. CL  (2014)

\bibitem{mikolov2010recurrent}
Mikolov, T.: Recurrent neural network based language model. (2010)

\bibitem{oh2000stochastic}
Oh, A.H., Rudnicky, A.I.: Stochastic language generation for spoken dialogue
  systems. In: Proc. NAACL. ACL (2000)

\bibitem{papineni2002bleu}
Papineni, K., Roukos, S., Ward, T., Zhu, W.J.: Bleu: a method for automatic
  evaluation of machine translation. In: Proc. ACL. pp. 311--318. ACL (2002)

\bibitem{pennington2014glove}
Pennington, J., Socher, R., Manning, C.D.: Glove: Global vectors for word
  representation. In: EMNLP. vol.~14 (2014)

\bibitem{ratnaparkhi2000trainable}
Ratnaparkhi, A.: Trainable methods for surface natural language generation. In:
  Proc. NAACL. ACL (2000)

\bibitem{rieser2010optimising}
Rieser, V., Lemon, O., Liu, X.: Optimising information presentation for spoken
  dialogue systems. In: Proc. ACL. pp. 1009--1018. ACL (2010)

\bibitem{stent2004trainable}
Stent, A., Prasad, R., Walker, M.: Trainable sentence planning for complex
  information presentation in spoken dialog systems. In: Proc. ACL. p.~79. ACL
  (2004)

\bibitem{tran2017natural}
Tran, V.K., Nguyen, L.M.: Natural language generation for spoken dialogue
  system using rnn encoder-decoder networks. In: CoNLL 2017. CoNLL 2017 (2017)

\bibitem{vinyals2015show}
Vinyals, O., Toshev, A., Bengio, S., Erhan, D.: Show and tell: A neural image
  caption generator. In: CVPR (2015)

\bibitem{wang2016inner}
Wang, B., Liu, K., Zhao, J.: Inner attention based recurrent neural networks
  for answer selection (2016)

\bibitem{thwsjy15}
Wen, T.H., Ga{\v{s}}i\'c, M., Kim, D., Mrk{\v{s}}i\'c, N., Su, P.H., Vandyke,
  D., Young, S.: {Stochastic Language Generation in Dialogue using Recurrent
  Neural Networks with Convolutional Sentence Reranking}. In: Proc. SIGDIAL.
  ACL (2015)

\bibitem{wen2016multi}
Wen, T.H., Gasic, M., Mrksic, N., Rojas-Barahona, L.M., Su, P.H., Vandyke, D.,
  Young, S.: Multi-domain neural network language generation for spoken
  dialogue systems. arXiv preprint arXiv:1603.01232  (2016)

\bibitem{wentoward}
Wen, T.H., Ga{\v{s}}ic, M., Mrk{\v{s}}ic, N., Rojas-Barahona, L.M., Su, P.H.,
  Vandyke, D., Young, S.: Toward multi-domain language generation using
  recurrent neural networks  (2016)

\bibitem{wensclstm15}
Wen, T.H., Ga{\v{s}}i\'c, M., Mrk{\v{s}}i\'c, N., Su, P.H., Vandyke, D., Young,
  S.: Semantically conditioned lstm-based natural language generation for
  spoken dialogue systems. In: Proc. EMNLP. ACL (2015)

\bibitem{wen2016network}
Wen, T.H., Vandyke, D., Mrksic, N., Gasic, M., Rojas-Barahona, L.M., Su, P.H.,
  Ultes, S., Young, S.: A network-based end-to-end trainable task-oriented
  dialogue system. arXiv preprint arXiv:1604.04562  (2016)

\bibitem{zhang2014chinese}
Zhang, X., Lapata, M.: Chinese poetry generation with recurrent neural
  networks. In: EMNLP. pp. 670--680 (2014)

\end{thebibliography}
\bibliographystyle{splncs03}

\end{document}